\documentclass{article}

\PassOptionsToPackage{numbers, compress}{natbib}
\usepackage[preprint]{arxiv}

\usepackage[utf8]{inputenc} % allow utf-8 input
\usepackage[T1]{fontenc}    % use 8-bit T1 fonts
\usepackage{hyperref}       % hyperlinks
\hypersetup{hidelinks}
\usepackage{url}            % simple URL typesetting
\usepackage{booktabs}       % professional-quality tables
\usepackage{amsfonts}       % blackboard math symbols
\usepackage{amsmath}        % align, binom, text, etc.
\usepackage{amssymb}        % checkmark and math symbols
\usepackage{algorithm}      % algorithm float
\usepackage{nicefrac}       % compact symbols for 1/2, etc.
\usepackage{microtype}      % microtypography
\usepackage{xcolor}         % colors
\usepackage{graphicx}       % resizebox, includegraphics
\usepackage{subcaption}    % subfigures
\usepackage{float}          % H placement for figures
\usepackage{pifont}         % ding symbols
\usepackage{multirow}       % multi-row table cells
\usepackage{colortbl}       % table cell background colors
\usepackage[most]{tcolorbox}
\usepackage{listings}
\usepackage{algpseudocode}
\usepackage{tabularx}
\usepackage{marvosym}       % \Letter envelope symbol
\usepackage{fontawesome5}   % GitHub icon
\definecolor{linkblue}{HTML}{1E88E5}
\definecolor{lightlinkblue}{HTML}{006FE6}
\definecolor{deepred}{HTML}{D21F1F}

\title{Claw-Anything: Benchmarking Always-On Personal Assistants with Broader Access to User's Digital World}

\author{%
  \hspace*{-1.5em}%
  \begin{tabular}{c}
    \textbf{Yusong Lin$^{1,2,\dagger}$ \quad Xinyuan Liang$^{2,3,\dagger}$ \quad Haiyang Wang$^{2,}$\textsuperscript{\Letter} \quad Qipeng Gu$^{2}$ \quad Siqi Cheng$^{2}$} \\
    \textbf{Jiangui Chen$^{2}$ \quad Shuzhe Wu$^{2}$ \quad Feiyang Pan$^{2}$ \quad Lue Fan$^{4}$ \quad Sanyuan Zhao$^{1,}$\textsuperscript{\Letter} \quad Dandan Tu$^{2,}$\textsuperscript{\Letter}} \\[4pt]
    \hspace*{-1.5em}{{\normalfont\small $^1$Beijing Institute of Technology \quad $^2$Huawei Technologies Co., Ltd}} \\
    \hspace*{-1.5em}{{\normalfont\small $^3$Peking University \quad $^4$Institute of Automation, Chinese Academy of Sciences}} \\[2pt]
    \hspace*{-1.5em}{{\normalfont\small \textsuperscript{\Letter}Corresponding authors \quad $^\dagger$Intern at Huawei~~}} \\[3pt]
    \hspace*{-1.5em}{\normalfont\small
      \faGithub\ Code: \href{https://github.com/LiberCoders/CLaw-Anything}{\textcolor{lightlinkblue}{\texttt{github.com/LiberCoders/Claw-Anything}}}
      \quad
      \raisebox{-0.3em}{\includegraphics[height=1.2em]{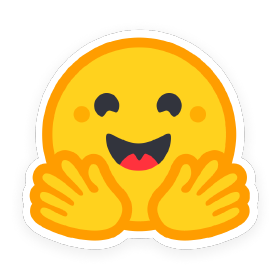}}
      Dataset: \href{https://huggingface.co/datasets/LiberCoders/Claw-Anything}{\textcolor{lightlinkblue}{\texttt{LiberCoders/Claw-Anything}}}
    } \\[4pt]
    \hspace*{-1em}{\normalfont\small\ttfamily \{linyusong4, haiyang.wang@huawei.com\}} \\[8pt]
    \hspace*{0.em}{\normalfont\small
      \textcolor{deepred}{\rule{0.14\textwidth}{0.45pt}}\quad
      \href{https://github.com/LiberCoders/CLaw-Anything/}{\textcolor{deepred}{\textbf{\textit{Scaling Agent Context: See Anything, then Do Anything.}}}}
      \quad\textcolor{deepred}{\rule{0.14\textwidth}{0.45pt}}
    } \\
  \end{tabular}
}

\begin{document}

\maketitle
\begin{figure}[h]
  \centering
  \vspace{-28pt}
  \includegraphics[width=0.98\linewidth]{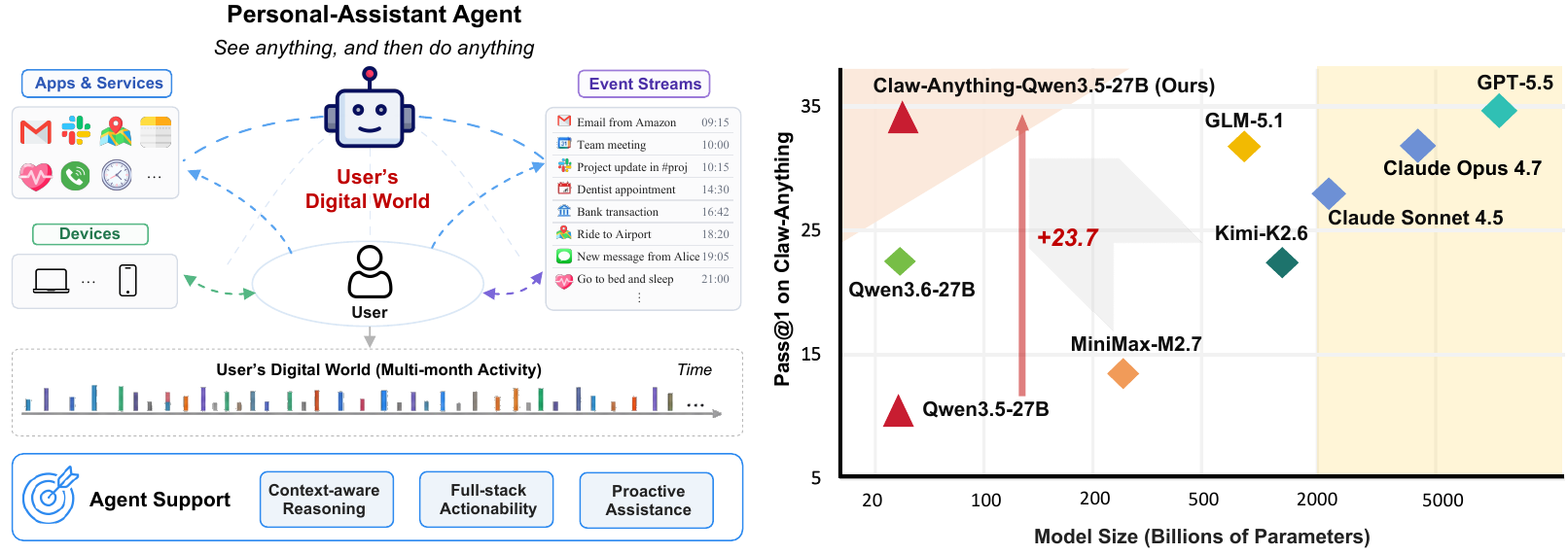}
  \vspace{-6pt}
  \caption{Overview of Claw-Anything and its empirical value. \textbf{Left:} Claw-Anything gives an always-on personal assistant broader access to the user's digital world, spanning services, devices, and long-horizon event streams, thereby expanding the range of tasks it can complete. \textbf{Right:} Enabled by our data pipeline, our model achieves the best pass@1 among open-weight models. The yellow region represents closed-source models, and the horizontal axis does not correspond to model sizes.}
  \label{fig:teaser}
\end{figure}

\begin{abstract}
Large language model agents are increasingly envisioned as always-on personal assistants with access to anything relevant in the user's digital world. Yet current systems operate over only narrow slices of that world, limiting context-sensitive reasoning and effective assistance. Existing benchmarks similarly provide only partial user state and therefore fail to capture performance in such a broad, always-on setting. To address this gap, we introduce Claw-Anything, a benchmark that expands agent context along three dimensions: long-horizon activity histories, interdependent backend services, and integrated GUI and CLI interaction across multiple devices. To instantiate this setting, we simulate months of user activity through multi-round event injection, producing complex world states and realistic noise, including irrelevant events and conflicting signals. Agents must reason over rich contextual environments while remaining robust to such noise. This expanded scope also enables the evaluation of proactive assistance, requiring agents to anticipate user needs and deliver timely recommendations. Experiments show that GPT-5.5 achieves only 34.5\% pass@1, substantially below prior benchmarks, underscoring a gap between current agent capabilities and the demands of always-on personal assistance. Alongside the benchmark, we release an automated data-generation pipeline that yields 2,000 training environments and improves the base model by 23.7\%, demonstrating its utility of scalable data infrastructure.
\end{abstract}

\section{Introduction}
\label{sec:intro}
Recent agent systems, such as the OpenClaw series~\citep{openclaw2026,nanobot2026,nanoclaw2026} and Hermes Agent~\citep{nous2026hermes}, are moving beyond one-shot task solving toward always-on personal assistance. Deployed within users' digital environments and equipped with long-term memory and background execution, these systems are expected to provide continuous, context-sensitive support over time. Yet user intent and activity are inherently distributed across heterogeneous digital artifacts, including historical events, backend services, and multiple devices. Effective assistance therefore requires broad access to the user's digital world, so that an agent can both perceive relevant state and act on it in a closed loop.

\begin{figure}[t]
  \centering
  \includegraphics[width=\linewidth]{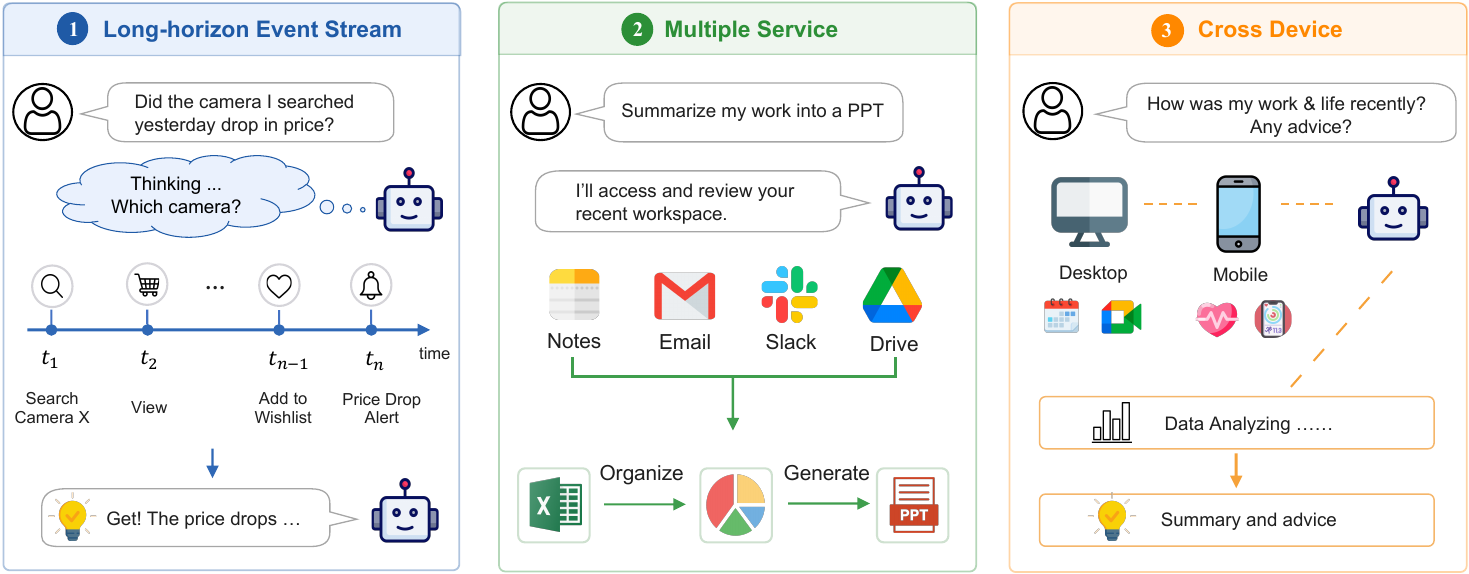}
  \caption{Three dimensions along which Claw-Anything expands agent context. \textbf{Left:} Long-horizon event streams provide a more complete view of the user's digital activity and support inference over evolving context. \textbf{Middle:} Access to multiple backend services enables cross-service coordination within a unified workflow. \textbf{Right:} Access across devices allows the agent to integrate distributed information and actions, broadening the range of tasks it can complete.}
  \label{fig:demo_of_three}
\end{figure}

Motivated by this shift, we argue that the effectiveness of personal assistants depends fundamentally on their operational scope: the set of digital states they can observe and the actions they can execute. As shown in Figure \ref{fig:teaser}, expanding this scope enlarges both the task space an agent can address and the context over which it can reason, enabling coordination across otherwise disconnected parts of the user's digital world. Similar patterns appear in other areas of AI: coding agents require access to the full codebase and executable environment to resolve realistic bugs~\citep{swebench,zhu2026swecontextbench,terminalbench2026}, while autonomous vehicles depend on broad sensor coverage for safe operation~\citep{sun2020scalability}. Consistent with this trend, recent systems increasingly expose richer digital interfaces to agents. Open-source projects such as CLI-Anything~\citep{cli-anything2026} and Gym-Anything~\citep{aggarwal2026gym}, as well as commercial platforms such as Google Workspace~\citep{gws-cli2026} and Feishu~\citep{lark-cli2026}, provide unified interfaces or programmable endpoints, making diverse software systems accessible to agents. These developments indicate that widening an agent's operational scope is critical for enabling it to perform complex tasks across the real-world digital environment.

However, current evaluation paradigms remain poorly aligned with this objective. Existing benchmarks~\citep{clawbench,wildclawbench,pinchbench,clawmark,qwenclawbench} typically expose only narrow, static slices of user state, omitting long-horizon activity, cross-service dependencies, and interaction across devices. As a result, they provide limited evidence about how agents perform when operating in richer, more realistic digital environments. To address this gap, we introduce Claw-Anything, a benchmark for evaluating personal-assistant agents under substantially broader access to the user's digital world.

As illustrated in Figure~\ref{fig:demo_of_three}, Claw-Anything expands agent context along three dimensions: i) long-horizon event streams that connect past and present through months of fine-grained activity records; ii) diverse, interdependent backend services spanning the principal digital spaces users inhabit; and iii) multiple devices with heterogeneous interfaces, including both GUI and CLI interaction. In this setting, the agent must integrate fragmented information and coordinate actions across time, services, and devices. The expanded context scope also enables evaluation of proactive assistance~\citep{ProAgentBench,yang2026contextagent}, requiring the agent to anticipate user needs and provide timely recommendations from context rather than merely react to explicit requests.

Constructing such environments at scale is challenging: it requires modeling extended time horizons, numerous services, and multiple devices while preserving realism and cross-component consistency. We therefore develop an automated pipeline that jointly synthesizes digital worlds and tasks. Starting from a minimal persona seed, an LLM-based simulator incrementally expands the user's digital world through multi-round event injection. At each step, it samples everyday events from a seed pool and updates both persistent world state and dynamic service traces, including sources such as email, calendars, and social platforms. Over time, the event history accumulates, the persona becomes more fully specified, and the environment acquires richer states and realistic noise, including irrelevant or contradictory events. Given the resulting digital world, the next event is instantiated as a persona-grounded task with an executable verifier, casting evaluation as completing the next step in an evolving digital life. Using this pipeline, we construct 200 human-verified evaluation tasks and 2,000 training environments, enabling Claw-Anything to function both as a benchmark and as scalable data infrastructure.

\begin{table}[t]
  \caption{Comparison of representative digital-agent benchmarks and Claw-Anything across three context-scaling dimensions, event streams, device interfaces, and services, plus proactivity. ``Event Stream'' denotes records of user activity in the digital environment; ``Device Interfaces'' the interaction surfaces in each task; ``Services'' the average and maximum number of services used per task; ``Context Length by words'' the length of textualized static states and dynamic event streams; and ``Proactive'' whether a task rewards action before an explicit user request.}
  \label{tab:rw-comparison}
  \centering
  \resizebox{\textwidth}{!}{%
 \begin{tabular}{lccccc|rr}
    \toprule
    Benchmark & \begin{tabular}[c]{@{}c@{}}Event \\ Stream\end{tabular} & \begin{tabular}[c]{@{}c@{}}Device \\ Interfaces\end{tabular} & \begin{tabular}[c]{@{}c@{}}\# Services \\(avg. / max.)\end{tabular} & Proactive & \begin{tabular}[c]{@{}c@{}}\# Context \\ Length by words \end{tabular} & \begin{tabular}[c]{@{}c@{}}\# Ins\\(Eval)\end{tabular} & \begin{tabular}[c]{@{}c@{}}\# Ins\\(Train)\end{tabular}\\
    \midrule
    ClawBench \citep{clawbench}         & \ding{55} & CLI & 1.6 ~(5)  & \ding{55} & ~~2.2k & 313 & 0\\
    WildClawBench \citep{wildclawbench} & \ding{55} & CLI & 0.5 ~(3)  & \ding{55} & ~~2.6k & 60 & 0\\
    PinchBench \citep{pinchbench}       & \ding{55} & CLI & 0.1 ~(3)  & \ding{55} & ~~1.7k & 53 & 0\\
    ClawMark \citep{clawmark}           & \ding{55} & CLI & 3.9 ~(5)  & \ding{55} & ~~2.0k & 100 & 0\\
    QwenClawBench \citep{qwenclawbench} & \ding{55} & CLI & 0.3 ~(6)  & \ding{55} & 12.1k & 100 & 0\\
    Claw-Eval \citep{claweval}          & \ding{55} & CLI & 1.3 ~(6) & \ding{55} & ~~5.3k & 300 & 0\\
    \midrule
    \textbf{Claw-Anything (ours)}       & \ding{51} & \textbf{CLI + GUI} & \textbf{10.1 ~(18)} & \ding{51} & ~\textbf{191.7k} & \textbf{200} & \textbf{2000}\\
    \bottomrule
  \end{tabular}}
\end{table}

Experiments reveal a substantial gap between current capabilities and the demands of full-access personal assistance. On Claw-Anything, GPT-5.5 achieves only 34.5\% on pass@1, substantially below performance reported on prior benchmarks. Several models that perform strongly on existing benchmarks also fail on ours, suggesting that Claw-Anything exposes failure modes underrepresented in prior evaluations and that current models remain unreliable even when given broader access to the user's digital world. Moreover, fine-tuning Qwen3.5-27B on 1,500 successful trajectories generated from the aforementioned training environments yields a 23.7\% improvement, indicating that Claw-Anything serves not only as a challenging benchmark but also as a practical source of scalable supervision.

In summary, our contributions are fourfold. 1) We identify the alignment between agent access and the user's digital world as a central challenge for personal-assistant agents, encompassing long-horizon event streams, interconnected services, and multi-device interaction. 2) We develop an automated pipeline for jointly simulating digital worlds and synthesizing tasks at scale, and use it to construct Claw-Anything, a benchmark of 200 human-verified task environments that expands agent context jointly along these dimensions while evaluating proactivity as a distinct capability, as shown in Table \ref{tab:rw-comparison}. 3) Through evaluation on Claw-Anything, we show that even GPT 5.5 attains only about 34.5\% success. 4) The same pipeline also yields 2,000 training environments, and fine-tuning Qwen3.5-27B on successful trajectories derived from them improves success by about 23.7\%, establishing Claw-Anything not only as a benchmark but also as a scalable data-generation pipeline.

\section{Related Work}
\label{sec:related}

\textbf{Benchmarks for Personal Assistant.} As claw-style agents have rapidly gained momentum, a growing family of benchmarks has emerged to measure their capabilities. ClawBench~\citep{clawbench} broadens coverage across a large set of standardized digital tasks, WildClawBench~\citep{wildclawbench} moves evaluation into more realistic open environments, PinchBench~\citep{pinchbench} centers on practical personal-productivity scenarios, ClawMark~\citep{clawmark} studies longer-horizon professional workflows, QwenClawBench~\citep{qwenclawbench} emphasizes execution in realistic user-distributed CLI tasks, and Claw-Eval~\citep{claweval} advances evaluation methodology through rubric-based assessment for open-ended trajectories. Collectively, these benchmarks have advanced the study of planning, tool use, and grounded interaction for digital agents. Yet they still largely cast the agent as a solver of localized tasks rather than an always-on assistant embedded in the user's broader digital world. Most remain confined to isolated, short-horizon, and relatively clean settings, offering limited traction on reasoning over noisy event streams, coordinating across devices and backend systems, or acting from accumulated personal context. To address this gap, Claw-Anything evaluates how agents perform when asked to operate over a much broader slice of the user's digital world, including long-horizon activity streams, interconnected systems, heterogeneous devices, and proactive opportunities.

\textbf{Scaling Agentic Training Environment.} In software-agent research, prior work on scalable environments has mainly followed two directions: code-centric scenaris~\citep{swebench,featurebench}, such as SWE-smith~\citep{swesmith} and SWE-Gym~\citep{swe-gym}; and terminal-centric scenarios~\citep{terminalbench2026}, such as CLI-Gym~\citep{cli-gym}, and TermiGen~\citep{termigen}. Together, these works suggest that scalable environments matter not only for evaluation, but also for broader agent development. This paradigm, however, remains underexplored in personal-assistant settings, where verifiable environment often depend on manual construction, limiting both realism and scalability. In this paper, we fill this gap by combining a realistic setting across services, time, and devices with a multi-round automated pipeline that jointly simulates personas, histories, and cross-service states. The resulting framework enables controlled variation in task difficulty and environmental complexity, providing a practical basis for scalable evaluation and development of personal-assistant agents.

\section{Methodolgy}
\label{sec:benchmark}

Claw-Anything is a benchmark for evaluating whether an agent can complete both reactive and proactive personal-assistant tasks when endowed with broad access to a user's digital world. Each task is grounded in a coherent persona and embedded in an environment spanning three contextual dimensions: long-horizon history, diverse backend services, and coordinated interactions across multiple devices with heterogeneous interfaces (e.g., GUI and CLI). Within this setting, the agent must isolate task-relevant signals from substantial background noise and execute required actions.
\begin{figure}[t]
  \centering
  \includegraphics[width=\linewidth]{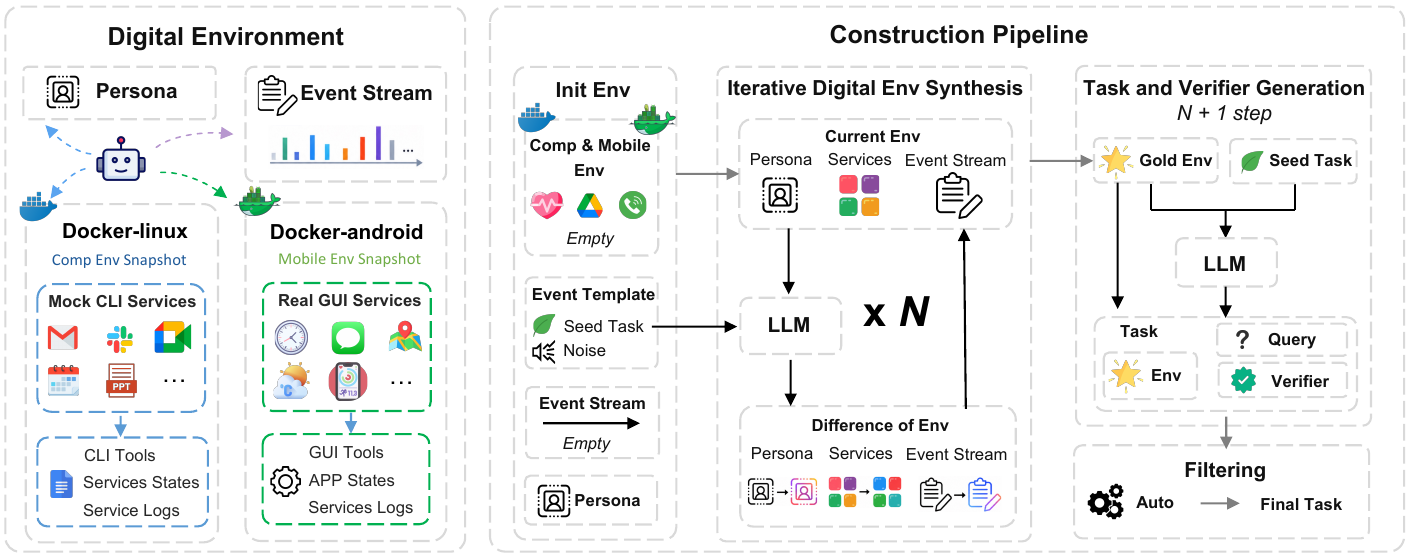}
  \caption{Claw-Anything environment and automated data pipeline. \textbf{Left:} The environment comprises connected devices with system event streams and multiple services with persistent states and service-specific histories. \textbf{Right:} From a persona-grounded initial state, the pipeline iteratively samples task or noise templates and uses an LLM-based simulator to adapt events and update the world state. A final simulation generates the task query, reference solution, and grader; automatic filtering then yields task instances, with optional human verification for benchmark cases.}
  \label{fig:pipeline}
\end{figure}
\subsection{Task Formulation}
\label{sec:task-formulation}
As illustrated in the left panel of Figure~\ref{fig:pipeline}, Claw-Anything first places the agent in a digital environment with access to as much of the user's digital world as possible, then formulates both reactive and proactive personal-assistant queries in this environment, and finally evaluates task completion with an executable verifier over the resulting interaction trace and task outcome.

\textbf{Context-rich digital environment.} We instantiate each task in a context-rich, realistic, and noisy digital environment. Formally, each environment is defined as $\mathcal{E}=(\mathcal{P},\mathcal{D},\mathcal{F},\mathcal{L})$, where $\mathcal{P}$ denotes a user persona specifying the user's profile and preferences; $\mathcal{D}$ denotes a set of devices with heterogeneous interfaces, including CLI-based computers and GUI-based mobile phones; $\mathcal{F}$ denotes a fixture bank of persistent states across more than forty backend services spanning lifestyle, work, and related domains; and $\mathcal{L}$ denotes a long-horizon activity stream covering over three months of system-level and service-specific logs. We further populate these environments with irrelevant events, services, and state to better approximate real-world settings, requiring agents to reason over large-scale context and complete tasks in a closed loop.

\textbf{Queries across time, services, and devices.} Each query is written in naturalistic and sometimes underspecified language, reflecting how users communicate in real personal-assistant settings. Solving these queries require the agent to identify task-relevant signals in the event stream and integrate information across services and devices, including CLI-based Linux Docker environments and GUI-based Android Docker environments. Beyond explicit requests, we also incorporate the heartbeat-style mechanism of OpenClaw, in which the agent periodically monitors the user's digital environment and produces contextually grounded recommendations without direct prompting.

\textbf{Outcome-oriented evaluation for multi-path tasks.} Our evaluation builds on the rubric-based framework of Claw-Eval~\citep{claweval}, combining rule-based checks with LLM judgments to produce both a soft score and a binary pass/fail label. Because many tasks admit multiple valid solution paths, we assign greater weight to the final outcome and correspondingly less to intermediate actions. This modification retains the strengths of rubric-based evaluation while better reflecting the open-ended nature of personal-assistant tasks.

\begin{algorithm}[t]
  \caption{Automated task generation pipeline.}
  \label{alg:algo}
  \small
  \textbf{Input:} seed persona $\mathcal{P}_0$; task-seed pool $\mathcal{S}$; noise-event pool $\mathcal{N}$; rollout horizon $R$; snapshot rounds $\mathcal{I}_{\mathrm{task}}$.\\
  \textbf{Initialize:} fixture state $\mathcal{F}\!\leftarrow\!\emptyset$, event log $\mathcal{L}\!\leftarrow\!\emptyset$, persona state $\mathcal{P}\!\leftarrow\!\mathcal{P}_0$, and task set $\mathcal{T}\!\leftarrow\!\emptyset$.\\
  \textbf{for} $r = 1,\dots,R$ \textbf{do}
  \begin{enumerate}
    \itemsep0.1em
    \item $e\!\leftarrow\!\mathrm{Sample}(\mathcal{S}, \mathcal{N}, \mathrm{noise\_ratio})$\hfill Sample a task or noise event
    \item $\tilde{e}\!\leftarrow\!\mathrm{AdaptToEnv}(e, \mathcal{P}, \mathcal{F}, \mathcal{L})$\hfill Ground it in the current environment
    \item Use an LLM to generate updates $\Delta\mathcal{F}, \Delta\mathcal{L}, \Delta\mathcal{P}$ from $\tilde{e}$
    \item Update the environment: $\mathcal{F}\!\leftarrow\!\mathcal{F} \cup \Delta\mathcal{F}$, $\mathcal{L}\!\leftarrow\!\mathcal{L} \cup \Delta\mathcal{L}$, $\mathcal{P}\!\leftarrow\!\mathcal{P} \cup \Delta\mathcal{P}$
    \item \textbf{if} $r \in \mathcal{I}_{\mathrm{task}}$ \textbf{then}
    \item[] \hspace*{2em}$X_r\!\leftarrow\!\mathrm{Snapshot}(\mathcal{F}, \mathcal{L}, \mathcal{P}, r)$\hfill Snapshot the current environment
    \item[] \hspace*{2em}$Q_r\!\leftarrow\!\mathrm{GenTaskQuery}(X_r)$\hfill Generate a task query
    \item[] \hspace*{2em}$V_r, A_{\mathrm{ref},r}\!\leftarrow\!\mathrm{GenVerifier}(Q_r, X_r)$\hfill Generate the verifier and reference answer
    \item[] \hspace*{2em}$\tau_r\!\leftarrow\!\mathrm{AutoFilter}(X_r, Q_r, V_r, A_{\mathrm{ref},r})$\hfill Filter the task instance
    \item[] \hspace*{2em}\textbf{if} $\tau_r \neq \varnothing$ \textbf{then} $\mathcal{T}\!\leftarrow\!\mathcal{T} \cup \{\tau_r\}$
  \end{enumerate}
  \textbf{Output:} Task set $\mathcal{T}.$ \hfill May undergo human verification for benchmark cases.\\
\end{algorithm}

\subsection{Construction Pipeline}
\label{sec:construction}

Manually constructing a context-rich digital world together with its associated tasks is prohibitively expensive and difficult to scale. We therefore generate both evaluation and training data with an automatic pipeline, illustrated in Algorithm~\ref{alg:algo} and Figure~\ref{fig:pipeline}, that incrementally builds an evolving user environment, extracts tasks from intermediate states, and removes low-quality instances.

\textbf{Stage I: Iterative digital environment synthesis.} We first construct an evolving digital environment through an iterative generation loop. At each round, the pipeline samples either a task template or a noise template from a predefined seed pool and conditions the LLM on the current persona and world state to generate the corresponding fixtures, event logs, and persona updates. Over multiple rounds, an initially sparse persona is transformed into a temporally coherent environment with accumulated event streams and richer cross-component dependencies, providing the substrate for subsequent task construction. 

\textbf{Stage II: Task and verifier generation.} We then derive tasks from designated rounds of the simulation. For each selected round, the pipeline captures the corresponding environment state and prompts the LLM on it to generate three coupled artifacts: a user query, an executable verifier, and a reference solution. Each task is thereby grounded in a specific temporal slice of the same evolving digital world, rather than synthesized from an isolated static state.

\textbf{Stage III: Automatic filtering.} Because the pipeline depends on LLM generation, automated quality control is necessary. We therefore combine rule-based checks with LLM-based filtering to remove invalid instances before human review. Rule-based checks target surface inconsistencies, such as references to nonexistent tools or services. LLM-based filtering then evaluates higher-level validity by using the environment state and reference solution to determine whether a task is solvable and whether its verifier is logically consistent with the specification.

\textbf{Stage IV: Human verification with execution support.} Finally, we perform human verification supplemented by execution-based validation. A strong agent is given the reference solution and asked to execute the task in the environment with the verifier. Successful execution indicates that the task admits at least one valid solution consistent with the intended logic, enabling human reviewers to focus on assessing the consistency among the query, environment, and verifier. Instances that fail execution are escalated for manual review to determine whether they should be revised or discarded.

\subsection{Claw-Anything}
\label{sec:bench-stats}

\begin{figure}[t]
  \centering
  \begin{minipage}[t]{0.6\linewidth}
    \vspace{0pt}
    \centering
    \scriptsize
    \setlength{\tabcolsep}{4pt}
    \renewcommand{\arraystretch}{1.15}
    \resizebox{0.98\linewidth}{!}{%
    \begin{tabular}{ll|c|>{\columncolor{gray!12}}c>{\columncolor{gray!12}}c}
      \toprule
      \multirow{2}{*}{\textbf{Category}} & \multirow{2}{*}{\textbf{Metric}} & \multirow{2}{*}{\textbf{Claw-Eval}} & \multicolumn{2}{c}{\textbf{Claw-Anything}} \\
      \cmidrule(lr){4-5}
      & & & Eval & Train \\
      \midrule
      \multirow{1}{*}{Size} & \# Instance & 300 & 200 & 2000 \\
      \midrule
      \multirow{2}{*}{Context Text} & \# Word of fixture & 5.3k & 108.0k & 97.3k \\
       & \# Word of log & 0 & 83.7k & 65.7k \\
      \midrule
      \multirow{2}{*}{Services} & \# Task-involved & 1.3 & 10.1 & 9.2 \\
      & \# Env-support & 19 & 35 & 35 \\
      \midrule
      Devices & Support Type & CLI & CLI + GUI & CLI + GUI \\
      \bottomrule
    \end{tabular}}
  \end{minipage}
  \begin{minipage}[t]{0.35\linewidth}
    \vspace{0pt}
    \centering
    \setlength{\fboxsep}{8pt}
    \includegraphics[width=\linewidth]{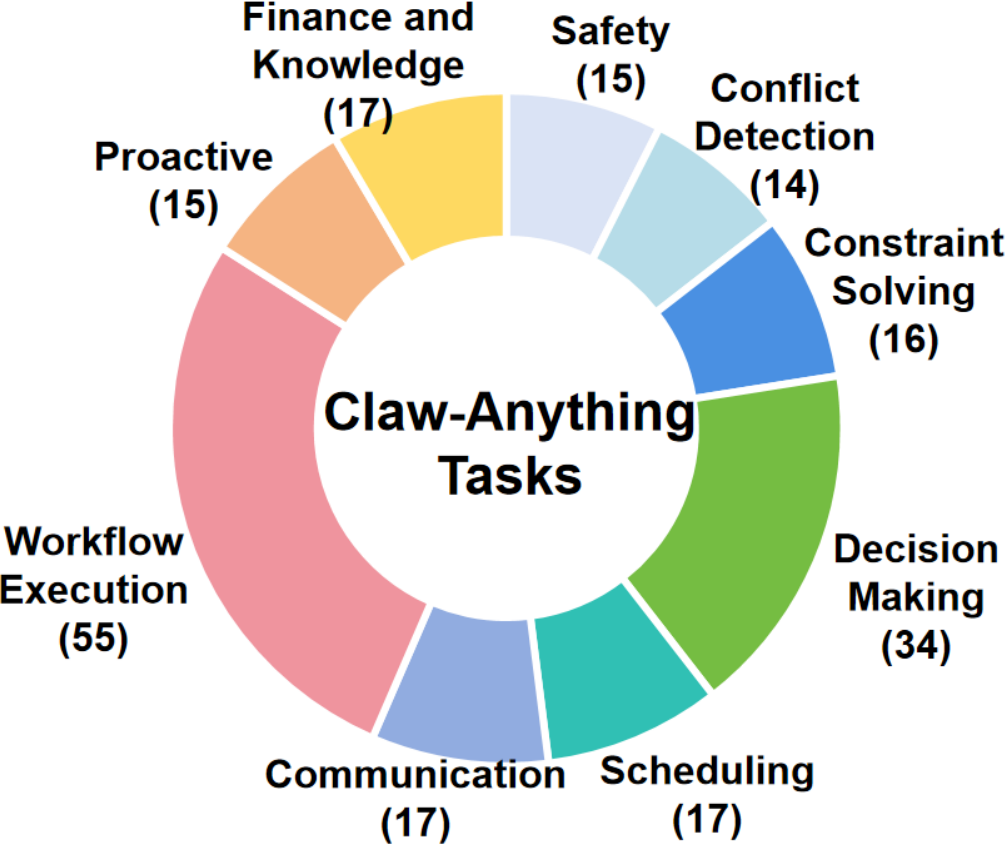}
  \end{minipage}
  \caption{Benchmark statistics of Claw-Anything. \textbf{Left:} Comparison with Claw-Eval in size, context length, services per task, and supported devices. \textbf{Right:} Category distribution of evaluation instances.}
  \label{tab:claweval-comparison}
\end{figure}

\textbf{Benchmark Statistics.} As shown in Figure~\ref{tab:claweval-comparison}, the full pipeline, including fourth-stage human verification, yields an evaluation set of 200 tasks, comprising 150 CLI-only tasks and 50 CLI+GUI tasks across 9 major categories. Compared with Claw-Eval, Claw-Anything provides a substantially richer perceptual context, with much longer temporal horizons, broader service coverage, denser cross-service dependencies, and task environments that require coordination across multiple devices.

\textbf{Trajectory Collection with Claw-Anything.} For training trajectory collection, we execute the first three stages of the automated pipeline to generate 2,000 task environments. To prevent contamination of the evaluation set, these environments are drawn from a persona pool fully disjoint from the evaluation personas. We then collect 1,500 successful trajectories from these environments for the subsequent post-training of Qwen3.5-27B.

\begin{table}[t]
  \caption{Main results on the Claw-Anything benchmark. We evaluate both state-of-the-art open- and closed-source models under a unified OpenHarness framework for fair comparison. The best result in each column is shown in bold.}
  \label{tab:main}
  \centering
  \small
  \resizebox{\linewidth}{!}{%
  \begin{tabular}{lrccc>{\columncolor{gray!20}}c|r}
    \toprule
    Model & \# Params & Score & Pass@1 & Pass@3 & Pass\textasciicircum{}3 & \# Tokens~(I~/~O) \\
    \midrule
    \textbf{\textit{Open-Source}} & & & & & & \\
    Qwen3.5-27B~\citep{qwen3p5}       & 27B & 0.50 & 9.8 & 19.0 & 2.0 & 83.8M~/~0.9M  \\
    MiniMax-M2.7~\citep{minimax_m27}   & 229B & 0.52 & 13.5 & 28.5 & 3.5 & 79.0M~/~1.1M\\
    Qwen3.6-27B~\citep{qwen3p6-27b}   & 27B & 0.58 & 22.5 & 42.0 & 6.0 & 99.4M~/~2.0M \\
    Kimi-K2.6~\citep{kimi26}           & 1.1T & 0.57 & 22.8 & 44.0 & 6.5 & 178.1M~/~2.3M\\
    GLM-5.1~\citep{GLM51}             & 754B & 0.59 & 31.7 & 47.0 & \textbf{17.0} & 125.0M~/~2.2M\\
    \textbf{Claw-Anything-Qwen3.5-27B (ours)} & 27B & \textbf{0.61} & \textbf{33.5} & \textbf{52.0} & 15.5 & 117.8M~/~1.1M\\
    \rowcolor{blue!10}\textit{Gain over Qwen3.5-27B} & - & \textit{+0.11} & \textit{+23.7} & \textit{+33.0} & \textit{+13.5} & - \\
    \midrule
    \textbf{\textit{Closed-Source}} & & & & & & \\
    Claude Sonnet 4.5~\citep{sonnet45} & - & 0.59 & 28.0 & 45.0 & 12.0 & 149.0M~/~1.5M\\
    Claude Opus 4.7~\citep{opus47}     & - & 0.62 & 31.8 & 48.0 & 13.5 & 123.5M~/~1.5M\\
    GPT-5.5~\citep{openai_gpt55}       & - & \textbf{0.65} & \textbf{34.5} & \textbf{53.5} & \textbf{20.0} & 77.7M~/~0.9M\\
    \bottomrule
  \end{tabular}}
\end{table}

\section{Experiment}
\label{sec:experiments}
\subsection{Main Results of Claw-Anything}
\textbf{Frontier baselines.} We benchmark a broad set of frontier LLMs, covering open-source families such as Qwen series~\citep{qwen3p5,qwen3p6-27b}, MiniMax 2.7~\citep{minimax_m27}, GLM 5.1~\citep{GLM51}, and Kimi 2.6~\citep{kimi26}, as well as closed-source models including Claude Opus 4.7~\citep{opus47} and GPT-5.5~\citep{openai_gpt55}. All models are evaluated under OpenHarness~\citep{openharness}, a widely adopted ultra-lightweight agent scaffold for personal agents implemented in pure Python. Following Claw-Eval, we use Claude Sonnet 4.5 as judge model and report Pass@1, Pass@3, and Pass\textasciicircum{}3 as the primary metrics, where Pass\textasciicircum{}3 requires success in all three independent runs. We further use continuous execution score and token consumption as complementary indicators of solution quality. Table~\ref{tab:main} summarizes the results. Even the strongest closed-source model reaches only 20.0\% on Pass\textasciicircum{}3, which suggests that bringing the agent's perceptual scope closer to that of the user materially increases benchmark difficulty, because success now depends on both accurate understanding of the user's digital environment and correct action grounded in that context.

\textbf{Improvement from collected training trajectories.} We further assess whether the automated pipeline serves not only as an evaluation infrastructure but also as a source of effective training data. Specifically, we construct 2,000 training tasks, collect 1,500 successful trajectories, and use them to fine-tune Qwen3.5-27B for 10 epochs. The resulting models improve over its base model by 23.7\% on pass@1, outperform all other open-source baselines on Claw-Anything, and reduce the gap to closed-source models. Figure~\ref{fig:scaling_law} further shows that performance increases steadily with the number of collected training trajectories. Together, these results indicate that data produced by our pipeline is effective for post-training and yields substantial gains on this benchmark.
\subsection{Ablation Study}
We conduct ablations on the key design choices of Claw-Anything, including scaling context in Section~\ref{sec:ab_scaling_context}, data pipeline in Section~\ref{sec:ab_data_pipeline}, and evaluation setting in Section~\ref{sec:ab_eval_setting}. Due to space constraints, additional experimental details are provided in the appendix.
\subsubsection{Scaling Context}
\label{sec:ab_scaling_context}
This section ablates whether expanding the agent's operational scope unlocks previously infeasible tasks, and whether larger context constitutes a fundamental bottleneck for current agents.

\begin{figure*}[t]
  \centering
  \begin{subfigure}[t]{0.33\textwidth}
    \centering
    \includegraphics[width=\linewidth]{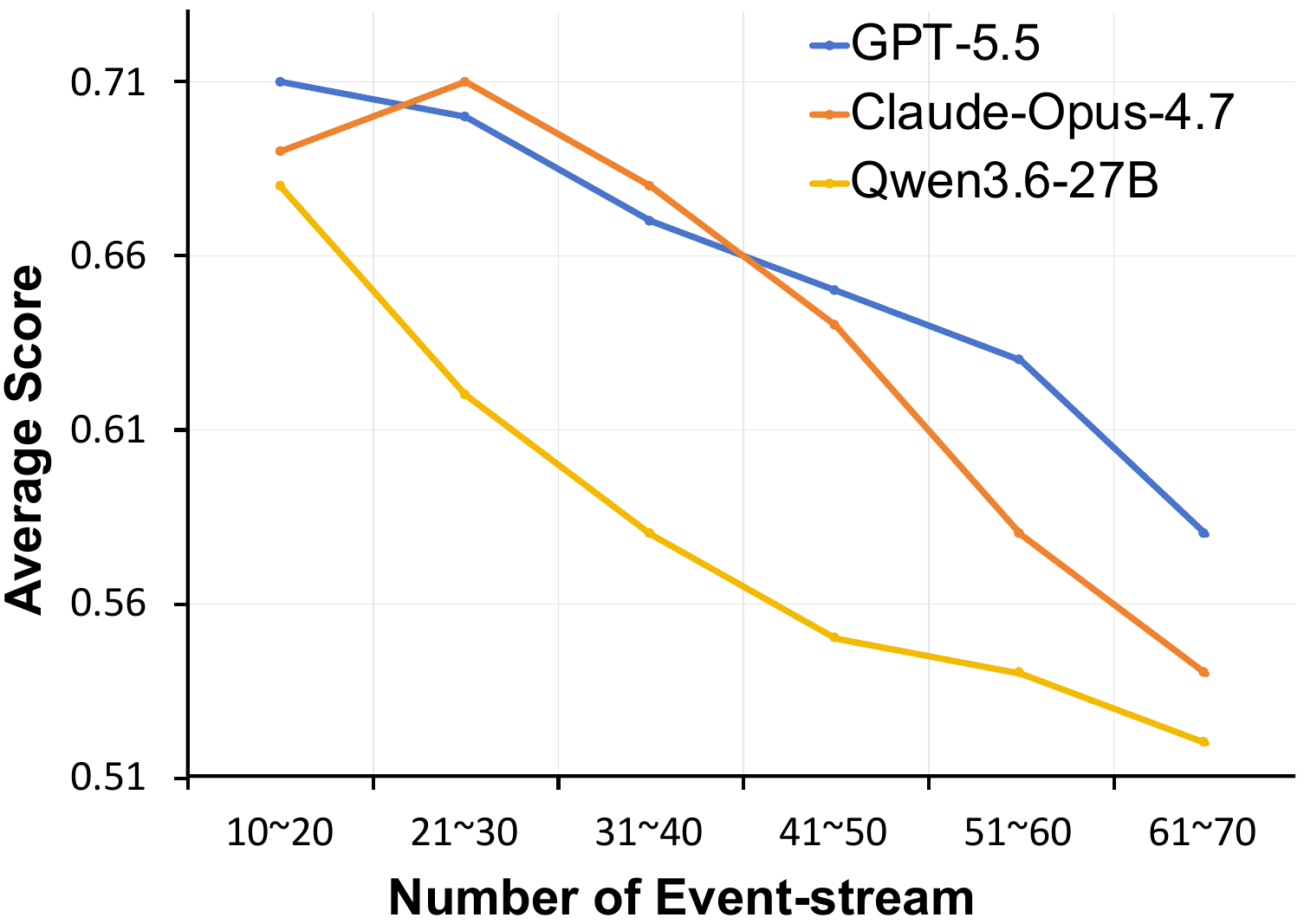}
    \caption{Long-horizon Event-stream.}
  \end{subfigure}\hfill
  \begin{subfigure}[t]{0.33\textwidth}
    \centering
    \includegraphics[width=\linewidth]{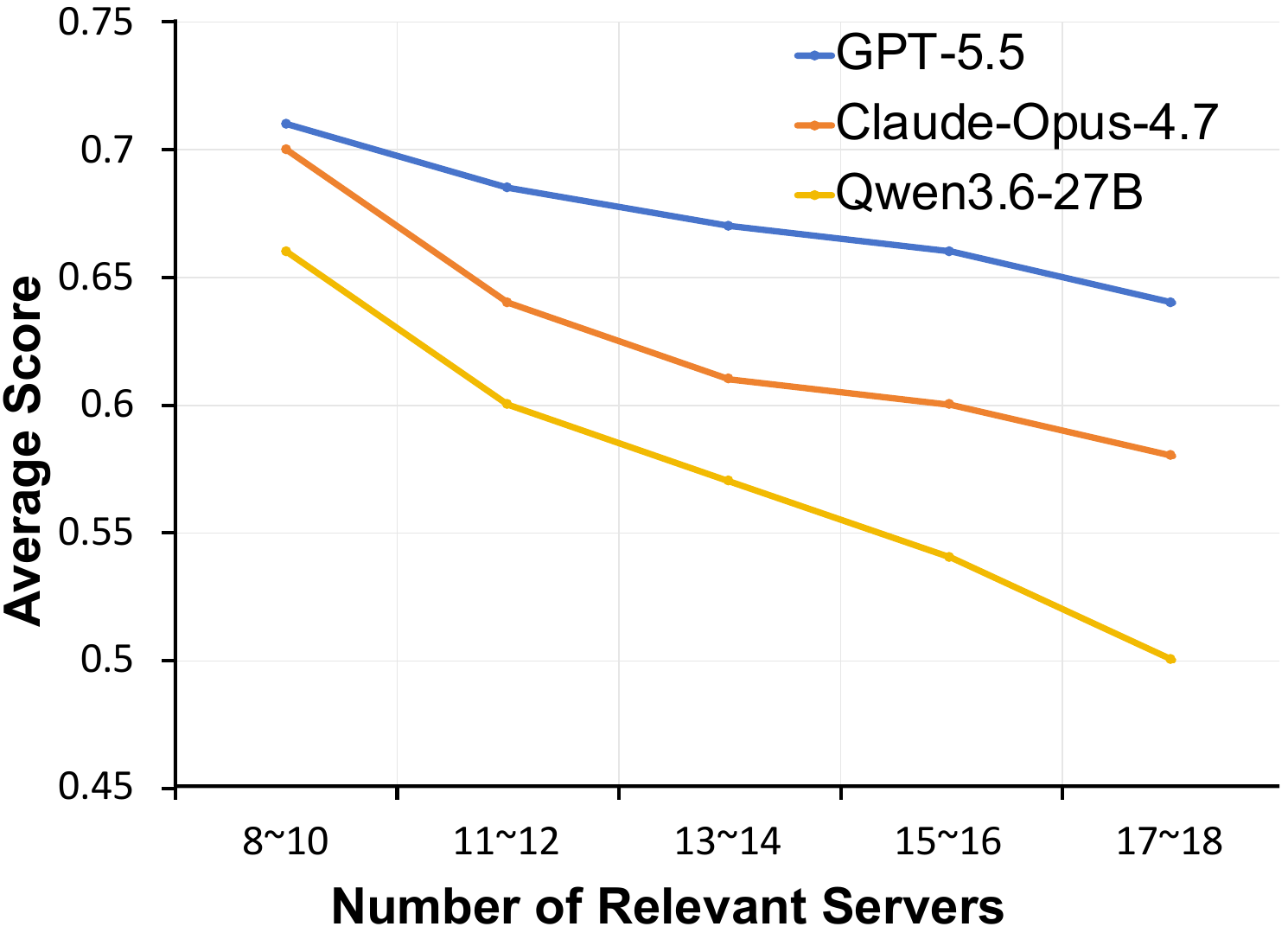}
    \caption{Multipile Backend Services.}
  \end{subfigure}\hfill
  \begin{subfigure}[t]{0.33\textwidth}
    \centering
    \includegraphics[width=\linewidth]{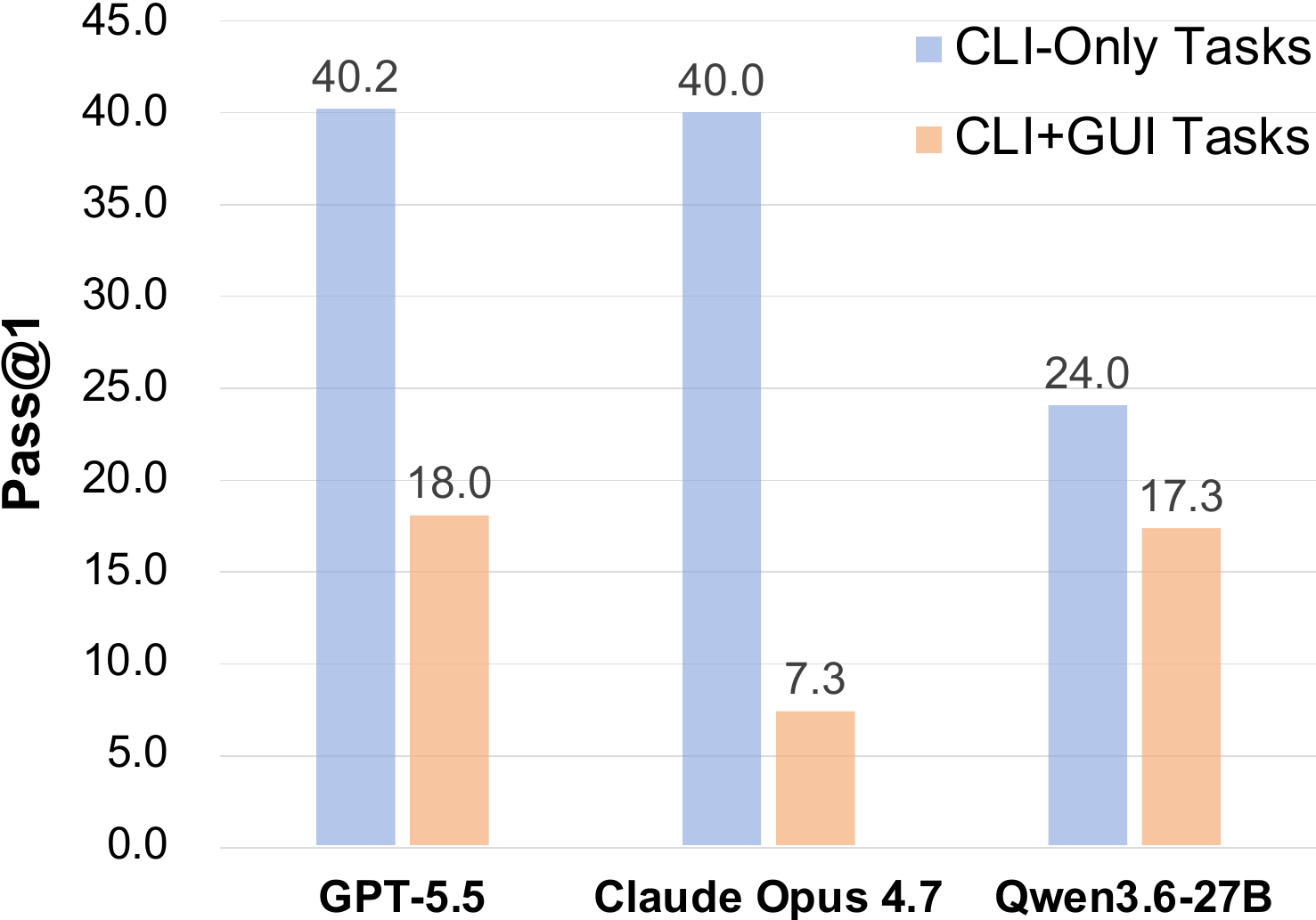}
    \caption{Cross-device (CLI~+~GUI).}
  \end{subfigure}
  \caption{Ablation of contextual scale, showing the effects of event-stream volume and the number of services on average score, as well as the effect of GUI access on Pass@1.
  }
  \label{fig:scaling}
\end{figure*}

\begin{figure*}[t]
  \centering
  \small
  \begin{minipage}[t]{0.48\textwidth}
    \vspace{0pt}
    \centering
    \captionof{table}{Effects of access to event streams, cross-service environments, and cross-device interaction on benchmark performance, together with a comparison between proactive and reactive tasks. All results are reported in Pass@1.}
    \label{tab:ablate_three_dim}
    \begin{tabular}{lcc}
      \toprule
      Factor & w/ & w/o \\
      \midrule
      Event Stream   & 21.0 & 0.0\\
      Cross-services & 24.0 & 0.0\\
      Cross-devices  & 16.0 & 2.0\\
      \midrule
      \midrule
      Task Type    & Reactive  & Proactive \\
      \midrule
      Pass@1    & 25.9 & 6.7 \\
      \bottomrule
    \end{tabular}
  \end{minipage}\hfill
  \begin{minipage}[t]{0.48\textwidth}
    \vspace{0pt}
    \centering
    \includegraphics[width=\linewidth]{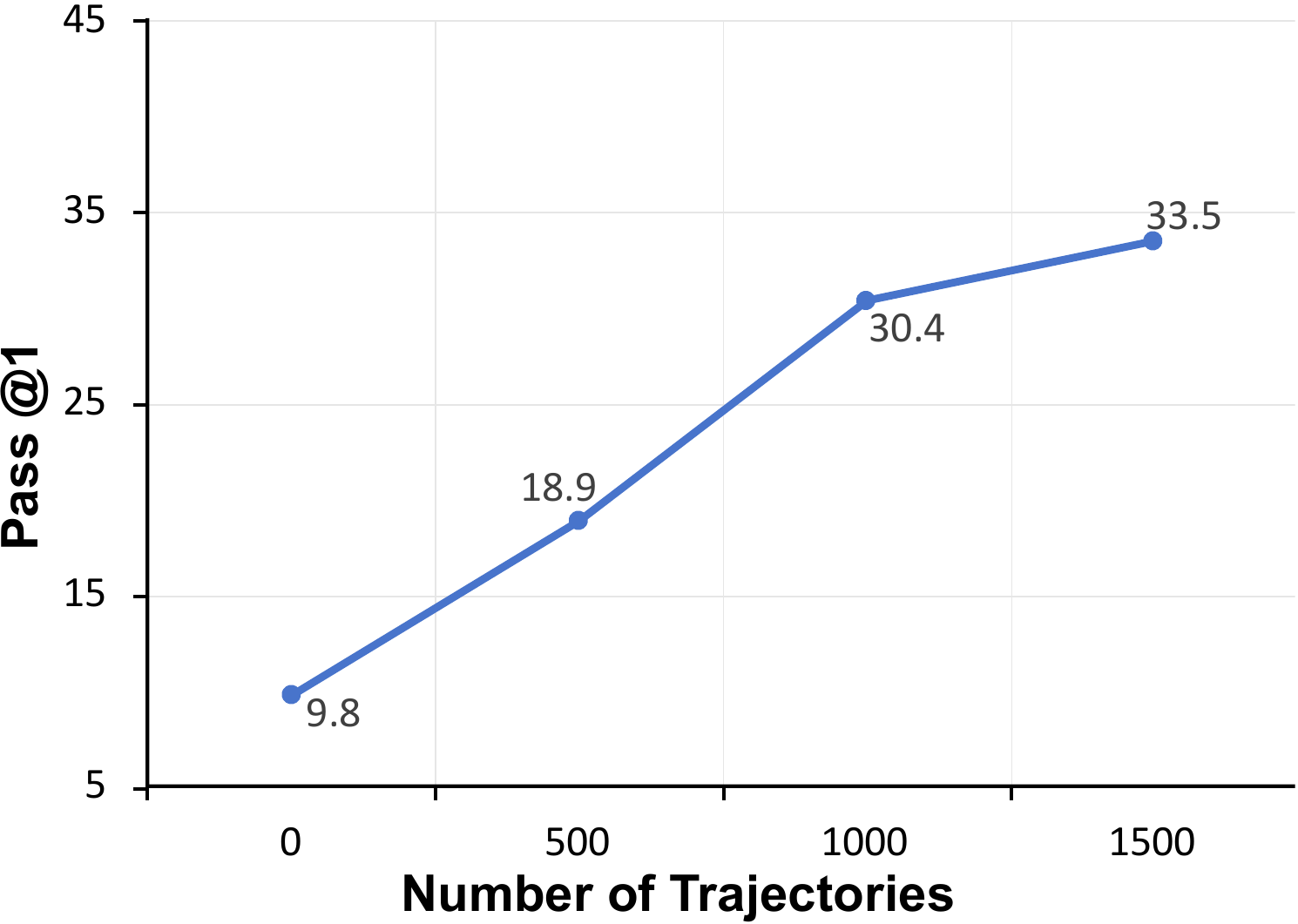}
    \captionof{figure}{Trajectory scaling.}
    \label{fig:scaling_law}
  \end{minipage}
\end{figure*}

\textbf{Long-horizon event streams.} We ablate both the availability of event streams and the length of history exposed to the agent. As shown in Table~\ref{tab:ablate_three_dim}, success rates drop substantially when event streams are removed, because many of these tasks inherently depend on information contained in the event history rather than in the static service fixtures alone. This finding supports our central claim that event streams enlarge the set of solvable tasks by extending the agent's operational scope toward that of the user. Figure~\ref{fig:scaling} further shows that, even when event streams are available, performance degrades as the history grows longer, suggesting that current models still struggle to effectively leverage long-horizon context despite having a broader field of view.

\textbf{Cross-backend services.} We ablate multi-service coordination by masking the tools required for tasks that span multiple backend services. As shown in Table~\ref{tab:ablate_three_dim}, success rates collapse to nearly zero once these tools are removed, indicating that many tasks intrinsically require the agent to retrieve information and execute actions across services rather than within a single isolated backend. This result underscores the importance of granting personal-assistant agents access to a digital ecosystem. Figure~\ref{fig:scaling} further shows that, even when all relevant tools are available, performance declines as the number of involved services increases. This trend suggests that cross-service coordination remains a major challenge for current models and a key target for future improvement.

\textbf{CLI--GUI collaboration.} We further ablate cross-interface coordination by removing GUI access and restricting the agent to CLI-only execution. As shown in Table~\ref{tab:ablate_three_dim}, tasks that intrinsically require CLI--GUI collaboration become nearly unsolvable in this setting, whereas restoring joint CLI+GUI access make them tractable again. At the same time, Figure~\ref{fig:scaling} shows that even with both interfaces available, performance on CLI--GUI collaborative tasks remains substantially below that on pure CLI tasks. Taken together, these results show that connecting CLI and GUI unlocks a new boundary of solvable task for agents, while robust coordination across heterogeneous interaction modalities remains a major challenge for current agent systems.

\begin{figure*}[t]
  \centering
  \begin{subfigure}[t]{0.33\textwidth}
    \centering
    \includegraphics[width=\linewidth]{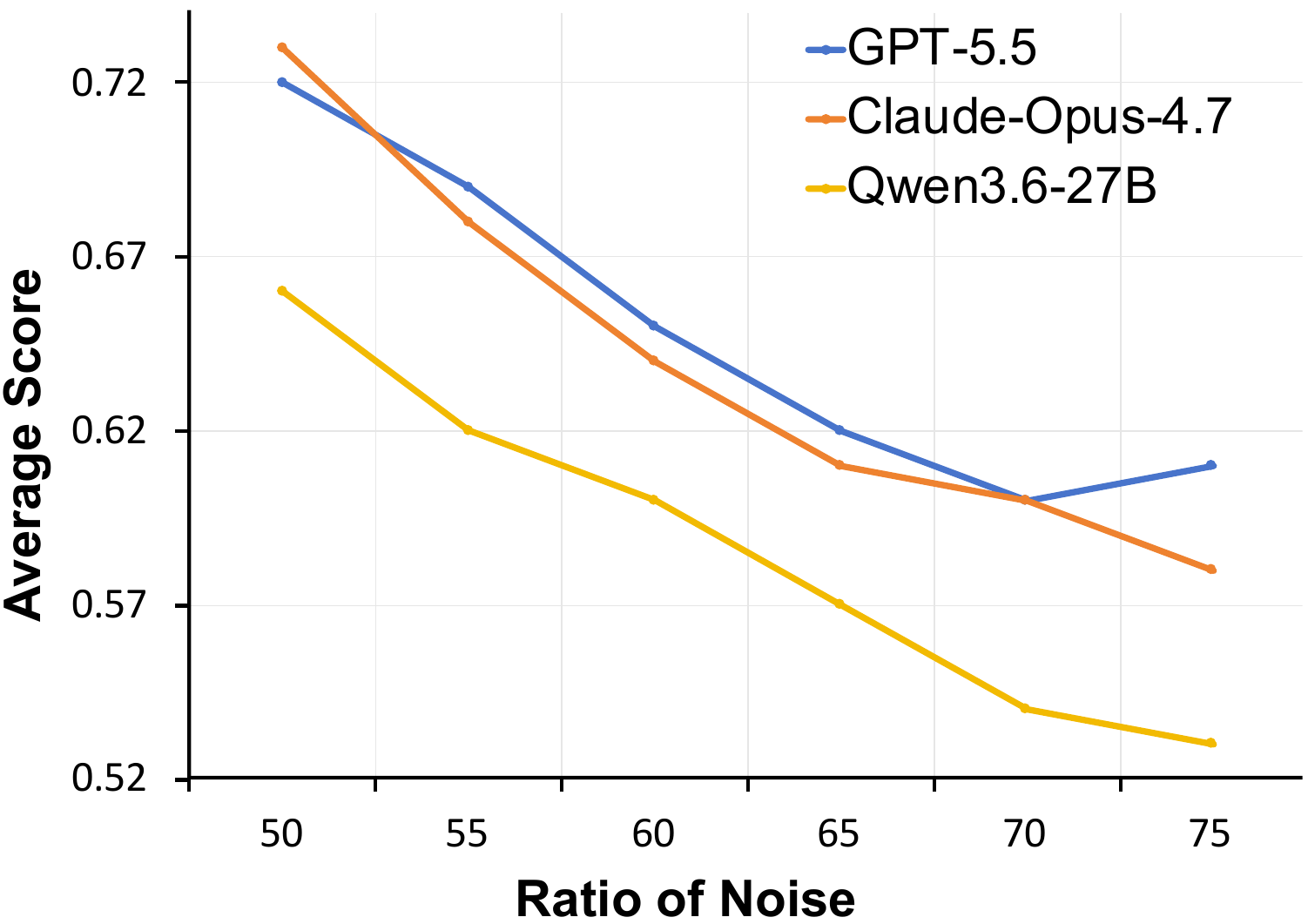}
    \caption{Ratio of noise rounds.}
  \end{subfigure}\hfill
  \begin{subfigure}[t]{0.33\textwidth}
    \centering
    \includegraphics[width=\linewidth]{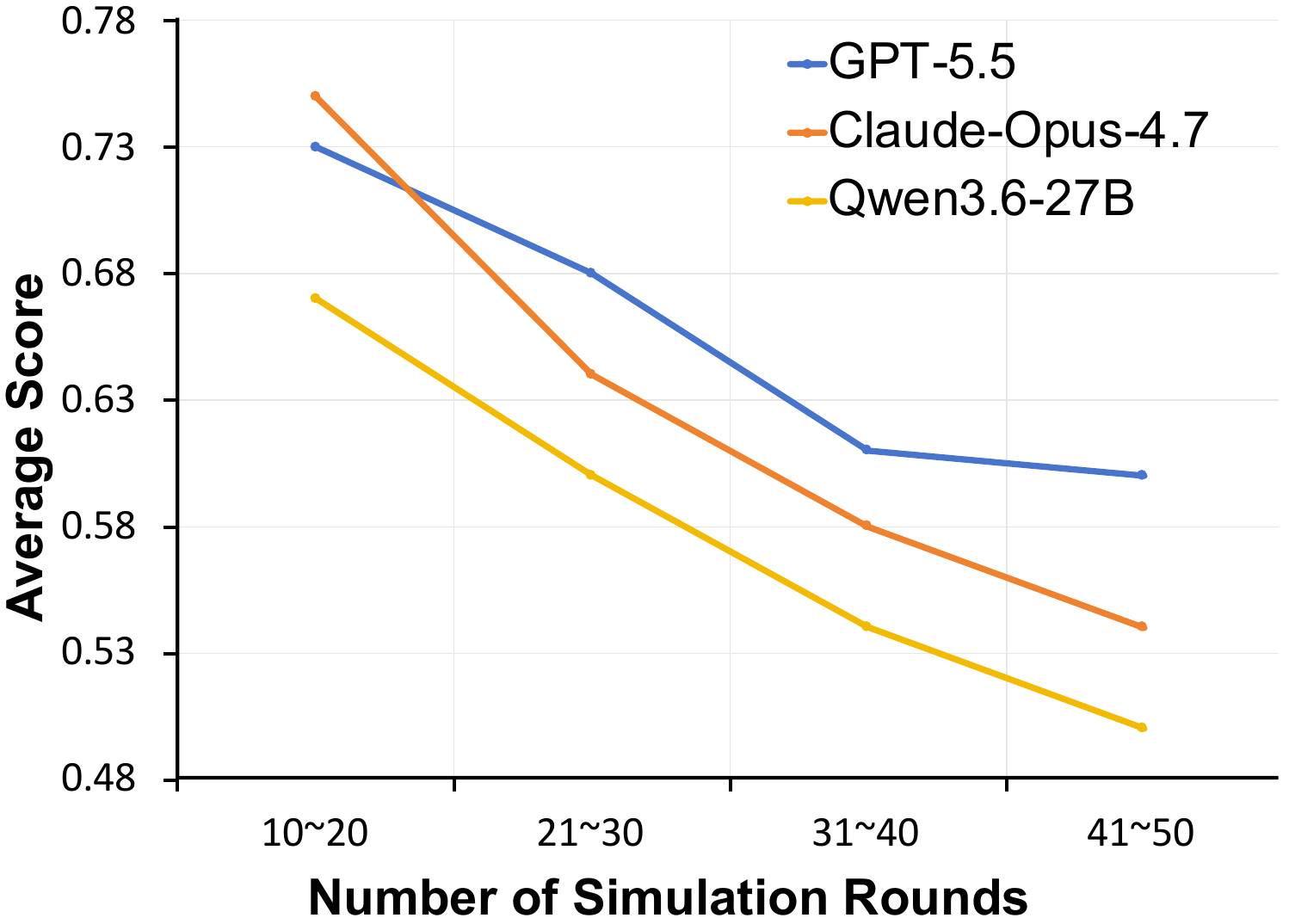}
    \caption{Simulation Rounds.}
  \end{subfigure}\hfill
  \begin{subfigure}[t]{0.33\textwidth}
    \centering
    \includegraphics[width=\linewidth]{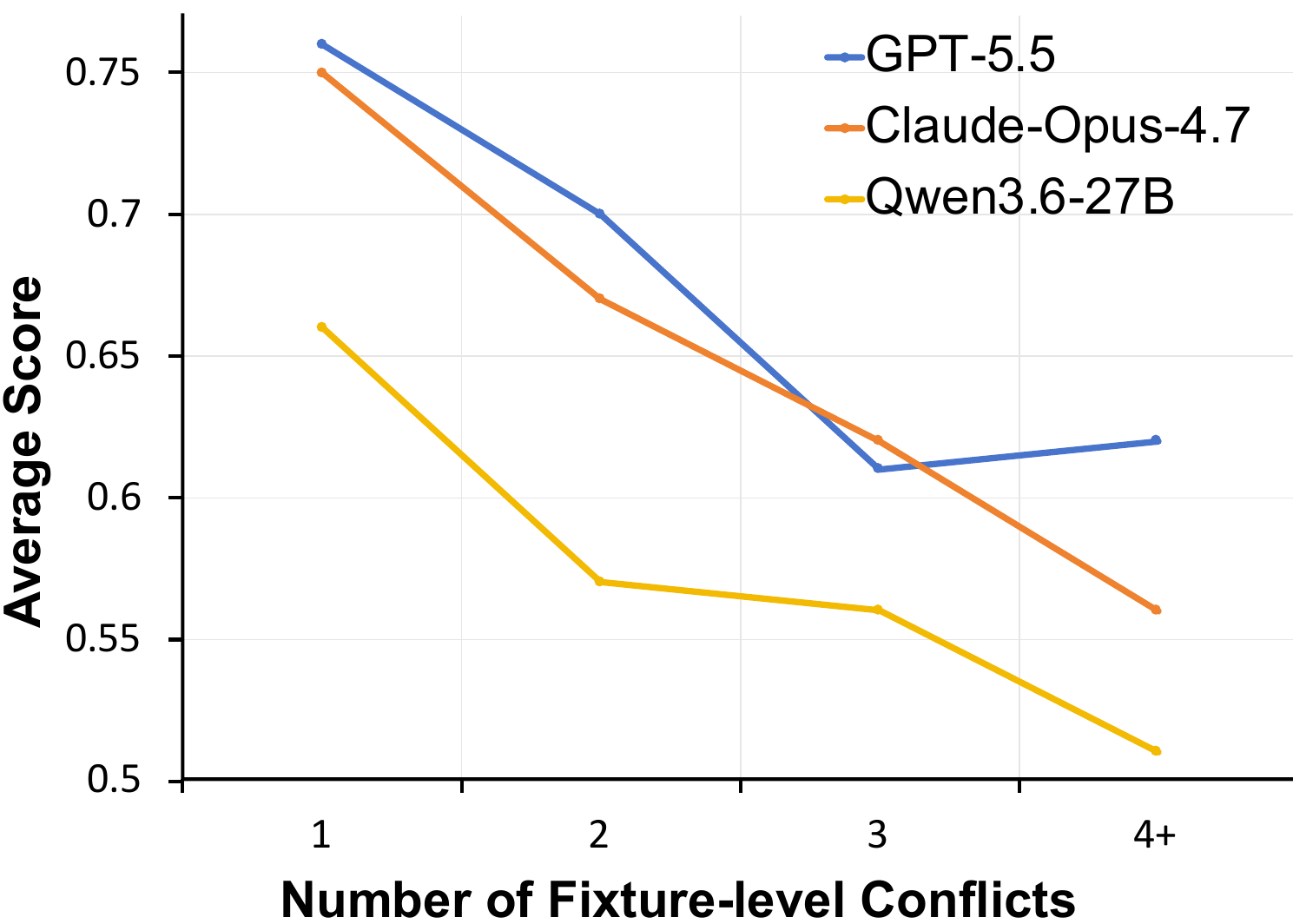}
    \caption{Fixture-level conflicts.}
  \end{subfigure}
  \caption{Ablation of the automatic data-generation pipeline, showing the effects of the noise-round ratio, the number of simulation rounds, and the number of fixture-level conflicts.}
  \label{fig:ablatedatapipeline}
\end{figure*}

\begin{figure*}[t]
  \centering
  \small
  \begin{minipage}[t]{0.48\textwidth}
    \vspace{0pt}
    \centering
    \captionof{table}{Skill-loading ablation. We compare full and lazy loading across models. Under lazy loading, the agent must select tools autonomously, making the setting much more challenging. All results are reported in Pass@1.}
    \label{tab:skill_loading}
    \begin{tabular}{lcc}
      \toprule
      Model & Full & Lazy \\
      \midrule
      Minimax 2.7       & 22.7 & 10.0 \\
      Qwen3.6-27B       & 24.7 & 23.7 \\
      GLM-5             & 29.3 & 14.0\\
      Claude Sonnet 4.6 & 43.0 & 26.7 \\
      \bottomrule
    \end{tabular}
  \end{minipage}\hfill
  \begin{minipage}[t]{0.48\textwidth}
    \vspace{0pt}
    \centering
    \includegraphics[width=\linewidth]{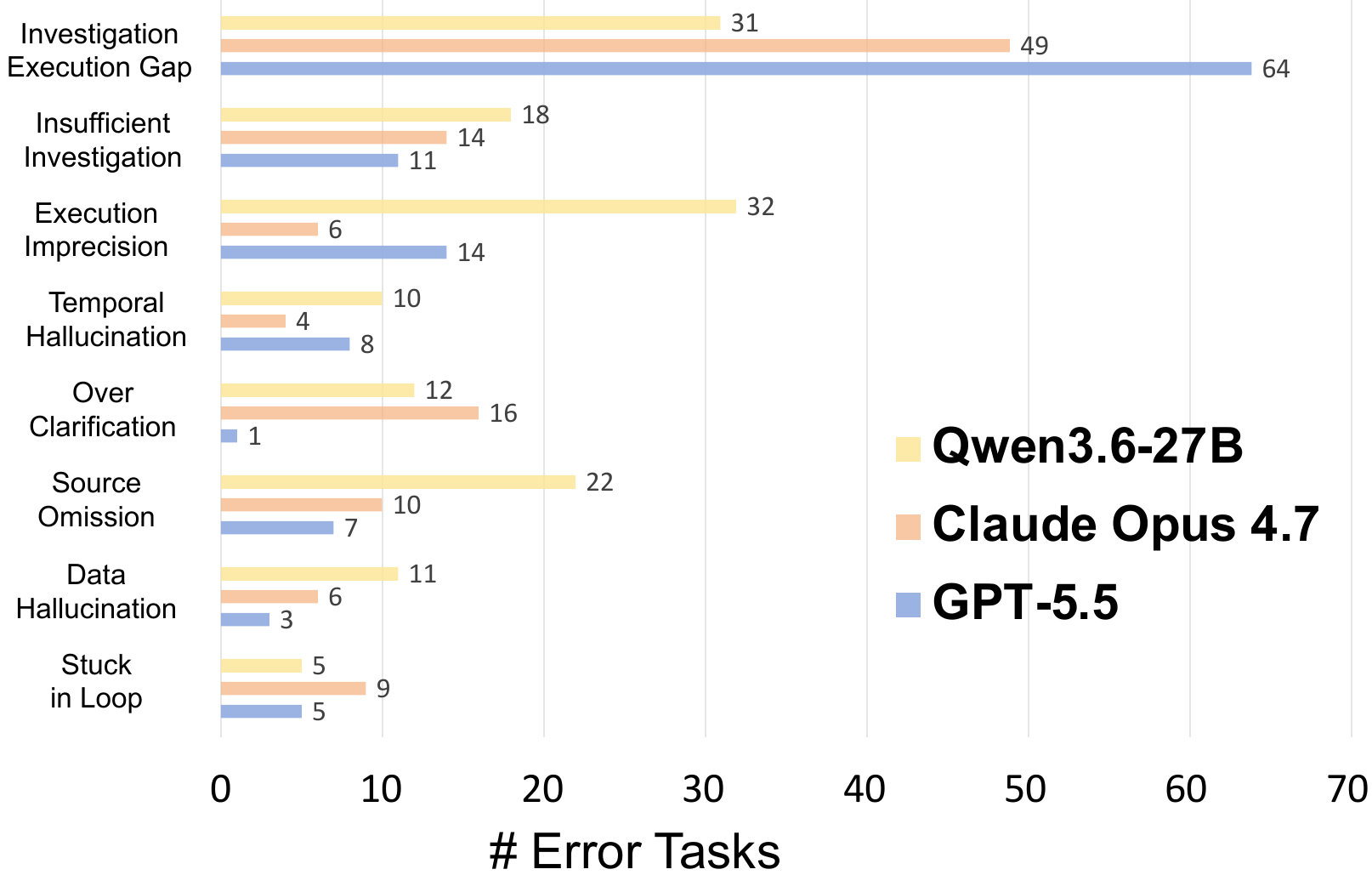}
    \captionof{figure}{Visualization of Failure modes.}
    \label{fig:failure_mode}
  \end{minipage}
  \vspace{-10pt}
\end{figure*}

\subsubsection{Data pipeline}
\label{sec:ab_data_pipeline}
\textbf{Noise injection ratio.}
Our generation pipeline injects a controllable amount of background noise into the user's digital environment. As shown in Figure~\ref{fig:ablatedatapipeline}, task success rates decline monotonically as the noise ratio increases, indicating that a denser stream of irrelevant events makes it harder for agents to recover the signals required for successful completion. This result suggests both that the pipeline can approximate more realistic, noisy environments and that environmental noise is itself a substantial source of difficulty.

\textbf{Persona richness.} In our data-generation pipeline, increasing the number of simulation rounds produces richer personas and more entangled task contexts. As shown in Figure~\ref{fig:ablatedatapipeline}, task success rates decrease steadily with the number of rounds, suggesting that persona richness is a key driver of benchmark difficulty. This trend further indicates that more fully developed personas yield more realistic and challenging evaluations for personal-assistant agents.

\textbf{Fixture-Level conflicts.} Our automated pipeline derives conflict-heavy tasks from seed tasks, forcing agents to reconcile inconsistent information across backend services. As shown in Figure~\ref{fig:ablatedatapipeline}, task success rates drop markedly as the level of conflict increases, indicating that cross-service inconsistency is a major source of difficulty. This result further supports the realism of the environment generated by our pipeline.

\subsubsection{Evaluation Setting}
\label{sec:ab_eval_setting}
\textbf{Proactivity.} As shown in Table~\ref{tab:ablate_three_dim}, proactive tasks are substantially more difficult than their reactive counterparts. This gap highlights proactivity as an important direction for future model development in personal-assistant agents. 

\textbf{Skill loading strategy.} We study how skill-loading strategy affects agent performance. Under full loading, the system prompt includes the complete specifications of all candidate tools. Under lazy loading, it provides only brief tool descriptions and a skill-loading utility, leaving the agent to determine which skills to use at test time. As shown in Table~\ref{tab:skill_loading}, lazy loading substantially degrades both success rate and stability. Qwen3.6-27B is relatively more robust in this setting, likely due to stronger recovery from incomplete skill context and more reliable tool selection.

\subsection{Faliure Mode Analysis}
Figure~\ref{fig:failure_mode} shows that the dominant failure mode across models is the investigation--execution gap. Agents often identify the relevant context yet fail to translate that understanding into successful action, indicating that execution remains the primary bottleneck in broad, always-on digital environments. Beyond this shared pattern, Qwen3.6-27B exhibits higher error rates of execution imprecision and source omission, whereas Claude Opus 4.7 more often over-clarifies or becomes trapped in loops. Hallucination-related errors are comparatively rare. 
\section{Conclusion}
\label{sec:conclusion}
We introduced Claw-Anything, a benchmark for evaluating personal-assistant agents under a substantially broader operational scope. By combining long-horizon event streams, diverse backend services, multi-device interaction, and proactive tasks, it captures core challenges that are largely absent from existing evaluations. Our results reveal a pronounced gap between current frontier models and the requirements of real-world assistance, with performance degrading as contextual breadth increases and proactive settings remaining especially difficult. The accompanying automated data-generation pipeline further supports scalable environment construction and provides a practical foundation for future research on personal-assistant agents.

\small
\bibliographystyle{plainnat}
\bibliography{references}

\newpage
\appendix

\section{Details of Task Generation Pipeline}

\subsection{Persona Creation and Enrichment}

In the generation of our Claw-Anything tasks, the persona plays a key role in guiding the environment evolving to make tasks personalized and diversified. At the stage I of our construction pipeline, i.e. iterative digital environment synthesis, an initial coarse persona is firstly generated, and then it is constantly enriched alongside sampling from a pool of tasks or events, which provide meta information for user event simulation.

The initial persona characterizes the basic information of a user, e.g., role and traits, as shown in Figure~\ref{fig:PersonaDemo}. It is simple and coarse, and can be easily generated by prompting any modern LLMs.

To guide the generation of more personalized, diversified and realistic tasks, the raw persona is elaborately enriched to include concrete user preferences and activities. The enrichment is driven by simulating a series of user events. Specifically, given a prepared pool of \textit{meta} tasks or events, characterizing conflicting or distracting factors, we constantly sample items from it. For each sampled item, a \textit{persona-specific} event is instantiated as the corresponding antecedents, consequences, and accompanying noises by prompting an LLM with the current persona and the sampled event information. Subsequently, the event instance is applied both to the current environment by injecting data into the App databases or logs and to the current persona by modifying the descriptions and adding activity records. After several rounds of update, the persona graduately deviates from the initialization and shows diversity in preferences and behavior patterns. An example of persona enrichment is shown in Figure~\ref{fig:PersonaEnrichDemo}. The prompt for instantiating a persona-specific event is present in Figure~\ref{fig:SeedTaskAdaptPrompt} and Figure~\ref{fig:DatabaseGeneration}, which are used to generate personalized task descriptions and the corresponding App data.

As for meta tasks, two types of them are introduced to characterize conflicting and distracting factors, respectively, aiming to make the task scenarios more realistic. For convenience of description, we call the former seed tasks and the latter noise events. The seed tasks present a certain pattern of conflicts that are commonly seen in the real world, e.g. time slot conflicts, information contradiction, financial limitation. Figure~\ref{fig:ConflictdescribingSeedTask} presents an example of the seed task with conflicts in the information provided. Such conflicts pose great challenges for resolving the target task, requiring intellectual thinking and appropriate handling of conflicts.

The noise events add distractions that are irrelevant with the target task to be resolved. We curate a library of routine activity patterns, e.g. scanning the inbox, drafting an email and discarding it, jotting a note and then deleting it, browsing the calendar or RSS feed, or checking an inventory dashboard. These activities can be realized as concrete dated sessions spread across the persona's working days. Some of them are purely \emph{ephemeral} and surface only in the activity logs, whereas others are \emph{trace-leaving} and additionally deposit residual records (such as deleted notes, discarded drafts, or cancelled todos) into the corresponding app databases. This further increases the complexity and realism of the environment without interfering the targeted task.

\subsection{Task Query and Verifier Generation}

At the Stage II of our construction pipeline, the targeted task to be resolved and the corresponding verifier are generated. We start from the persona produced at the Stage I, randomly sample a conflict-describing seed task, and adapt it to the current persona to instantiate a persona-specific event as described above. An LLM is prompted to generate the antecedents that trigger this event and to articulate the problem that the agent is expected to solve. Finally, from a god's-eye view, i.e., with full visibility into the entire environment, a verifier or grader together with a reference solution is generated for scoring the agent's outcome at evaluation time. It is worth noting that the LLM responsible for producing the grader is supplied only with purposeful, noise-free information that is synthesized jointly with the task itself; the grader therefore enjoys a substantial information advantage over the agent under evaluation, which in turn underwrites the reliability of its judgments. 

\subsection{Task Validation}

To validate the generated task, we feed the task and its reference solution to an agent that executes them end-to-end in a simulated environment, thereby verifying that the task is both solvable and well-formed under the grader's recommended solution path.

\section{Claw-Anything Evaluation}

At evaluation time, we make targeted modifications to the agent's system prompt and tool interface, build a service that simulates the app backend environment to satisfy the agent's runtime needs, and adopt a tailored scoring scheme. We describe each of these in turn.

\subsection{Modifications to the System Prompt and Tools}

Every task is generated with a fixed internal ``current date'', so allowing the agent to determine the date on its own would cause results to drift across evaluation dates. We therefore inject the task-specific date directly into the system prompt, which guarantees that repeated evaluations of the same task remain comparable across different calendar dates.

We also modify the agent's tool interface so that, at evaluation time, the agent can access the app databases populated during task generation. Our evaluation supports two modes: \emph{skill mode} and \emph{tool mode}. In skill mode, the agent is provided with a single meta-tool for retrieving full tool specifications; the system prompt lists only the names and short descriptions of the available tools, and the agent must invoke the meta-tool to obtain a tool's full usage specification before calling it. In tool mode, by contrast, the names and full usage specifications of all tools are placed directly in the system prompt, allowing the agent to invoke them without any additional retrieval step. App log information is handled separately: we inject the app logs as files into the Docker environment in which the task is executed, declare their file paths in the system prompt, and rely on the agent's native file-system read tool to access them. Figure~\ref{fig:SystemPromptDemo} shows a system prompt using OpenHarness.

\subsection{Simulated Service Backend}

To support agent evaluation, we build a service that simulates the app backend environment~\ref{fig:AppBackendDemo}. This service parses the tool calls issued by the agent and either returns the requested information or performs write/delete operations, thereby providing a faithful execution of the agent's interactions with the apps. The service takes an agent tool call as input and returns either the requested data or the result of the corresponding interaction, which together give the agent full access to the app databases during evaluation.

\subsection{Scoring}

When computing the pass rate, we want failed tasks to also receive a meaningful continuous score. Our grader therefore additionally scores the agent's solution trajectory. Unlike tasks with a single canonical solution path, our tasks contain a large amount of noise---both semantically irrelevant noise and intentionally distracting noise---together with multiple alternative cues that reveal the persona's traits, as well as the foreshadowing and multi-threaded clues planted during task generation. As a consequence, our tasks are largely \emph{multi-path}: multiple solution trajectories can lead to the same correct answer. Scoring against a single canonical trajectory would therefore allow an agent that arrived at the correct answer via an alternative path to nevertheless fall below the pass threshold for lack of process credit. To avoid this pitfall, we adopt a \emph{decisive}, outcome-dominated scoring scheme: a correct outcome yields a score that by itself exceeds the pass threshold, whereas an incorrect outcome yields a score that falls far below it. This guarantees that any agent reaching the correct answer---regardless of the path taken---is credited with a passing score. Even when a task is not completed, the agent still receives a corresponding process score, enabling a more fine-grained evaluation of its behavior.

\subsection{Detailed Evaluation Setting and Result}

Claw-Anything benchmark consists of 150 CLI tasks and 50 CLI+GUI tasks. The 150 CLI tasks consist of 100 skill mode tasks and 50 tool mode tasks.
Detailed model performance on 150 CLI tasks and 50 CLI+GUI tasks are shown in Table~\ref{tab:guiper}.

\begin{table}[t]
    \centering
    \caption{Model performance on 150 CLI-only tasks and 50 CLI+GUI tasks(pass@1).}
    \label{tab:guiper}
    \begin{tabularx}{0.65\columnwidth}{lcc}
    \toprule
    Model & CLI-only tasks & CLI+GUI tasks \\
    \midrule
    Qwen3.5-27B             & 10.0 & 9.3 \\
    Qwen3.6-27B             & 24.0 & 18.0 \\
    MiniMax-M2.7            & 14.2 & 11.3 \\
    Kimi-K2.6               & 27.3 & 9.3 \\
    GLM-5.1                 & 36.9 & 16.0 \\
    Claw-Anything-Qwen3.5-27B & 38.4 & 18.7 \\
    Claude Sonnet 4.5       & 35.1 & 6.0 \\
    Claude Opus 4.7         & 40.0 & 7.3 \\
    GPT-5.5                 & 40.2 & 17.3 \\
    \bottomrule
    \end{tabularx}
\end{table}

\section{Training Details}

The learning rate is initialized at $2 \times 10^{-5}$ and follows a cosine decay schedule. To stabilize the early stage of training, we employ a linear warmup strategy for the first 5\% of the total training steps, during which the learning rate increases linearly from a minimum $1 \times 10^{-6}$ to $2 \times 10^{-5}$. We train models for 10 epochs. The batch size is set to 16. We adopt qwen3-coder as the agent template. The maximum sequence length is 100k tokens. 

\section{Limitations}
\label{sec:limitations}
Claw-Anything is a step toward evaluating always-on personal assistants with broader access to the user's digital world, but it still has important limitations. First, although the benchmark includes multiple backend services and cross-service dependencies, many of these services are still implemented as controllable mock environments rather than fully real-world systems. Second, our current setting still covers only a limited subset of the devices that shape real personal-assistant usage. While Claw-Anything already incorporates cross-device interaction, the connected device ecosystem remains incomplete relative to everyday settings in which users move fluidly among phones, laptops, tablets, wearables, smart-home devices, and other ambient computing endpoints.

\section{Social Impact}
\label{sec:social_impact}
Claw-Anything has the potential to create a positive social impact by supporting research on more capable and context-aware personal-assistant agents. Broader access to a user's digital world may enable systems that reduce coordination burden, help users manage complex information flows, and provide more timely assistance in everyday and professional settings. In particular, improved evaluation of long-horizon, cross-service, and cross-device reasoning can help the community better understand the current limitations of such systems and develop safer, more useful assistants.

At the same time, the capabilities studied in this benchmark also raise meaningful societal risks. Always-on assistants with broad digital access may amplify privacy concerns, since systems that can observe and act across multiple services and devices could expose sensitive personal information or enable overly intrusive behavior if deployed without sufficient safeguards. Relatedly, stronger autonomy and broader operational scope may increase the risk of erroneous actions, overreach, or misuse in high-stakes settings. Although our benchmark is designed for evaluation rather than direct deployment, we hope it encourages future work on safeguards such as permission boundaries, transparency, auditability, and user control, as well as on privacy-conscious designs that better align capable personal assistants with user interests and social expectations.

\begin{figure}[htbp]
\centering
\begin{tcolorbox}[
    colback=white,           
    colframe=black!60,         
    coltitle=white,         
    colbacktitle=black!60,
    title=Persona,    
    fonttitle=\bfseries,      
    boxrule=0.7pt,             
    arc=1mm,                   
    width=1\linewidth,      
]
\begin{lstlisting}[basicstyle=\footnotesize\ttfamily, breaklines=true]
persona_id: T14
persona_name: Abdullah Rashid
language: en
role: senior project engineer on mega-infrastructure builds
company: Dar Al-Handasah (seconded to a Neom district-level consortium)
industry: construction / civil engineering
seniority: senior
traits:
- 28 years across bridges, ports, and high-speed rail, most recently on Saudi giga-project packages
- manages a 40-person multinational field team across Arabic, English, and broken Mandarin with subcontractors
- devout but private about it, steps away for prayer without announcement and expects no accommodation beyond the schedule
- MBTI is ESTJ
- refuses to sign off on any RFI response he has not personally walked through on site, even when the schedule screams
- family lives in Amman; he rotates 6-weeks-on / 2-weeks-off and guards the off-rotation time fiercely
- keeps a worn leather notebook of lessons-learned from every project, numbered sequentially since 1998

\end{lstlisting}

\end{tcolorbox}
\caption{Example of an initial persona}
\label{fig:PersonaDemo}
\end{figure}

\begin{figure}[htbp]
\centering
\begin{tcolorbox}[
    colback=white,           
    colframe=black!60,          
    coltitle=white,            
    colbacktitle=black!60,
    title=Persona Enrichment,    
    fonttitle=\bfseries,       
    boxrule=0.7pt,             
    arc=1mm,                   
    width=1\linewidth,      
]
\begin{lstlisting}[basicstyle=\footnotesize\ttfamily, breaklines=true]
- name: Neom Coastal Connector Bridge Construction Schedule Optimization
  description: Series of time-cost trade-off decisions during the bridge construction phase, involving coordination between concrete works, steel erection sequencing, and subcontractor mobilization windows. Includes interactions with Chinese steel fabricators, local concrete suppliers, and consortium commercial controls. Documents Abdullah's approach to balancing liquidated damages risk against cost overruns while maintaining his personal verification standards.
  involved_services:
  - calendar
  - finance
  - kb
  - notes
  - contacts
  involved_records:
    gmail:
    - MSG-5001
    - MSG-5002
    calendar:
    - EVT-301
    - EVT-304
    contacts:
    - CON-201
    - CON-202
    todo:
    - TODO-501
    - TODO-502
    kb:
    - KB-401
    - KB-402
    - KB-403
    - KB-404
    finance:
    - TXN-6001
    - TXN-6003
    notes:
    - NOTE-101
    - NOTE-102
  signal_density: 0.0
  difficulty: hard
  category: decision_making
  thread_type: standard
  patrol_signals: []
\end{lstlisting}

\end{tcolorbox}
\caption{Example of persona enrichment}
\label{fig:PersonaEnrichDemo}
\end{figure}

\begin{figure}[htbp]
\centering
\begin{tcolorbox}[
    colback=white,           
    colframe=black!60,          
    coltitle=white,            
    colbacktitle=black!60,
    title=Task with Conflicting Factors,    
    fonttitle=\bfseries,       
    boxrule=0.7pt,             
    arc=1mm,                   
    width=1\linewidth,      
]
\begin{lstlisting}[basicstyle=\footnotesize\ttfamily, breaklines=true]
seed_id: M000
name: Hidden information conflict
category: decision_making
task_content: conflict_detection
interaction_mode: standard
plausible_content_directions: [conflict_detection, scheduling, workflow_execution, info_triage, risk_safety]
difficulty: hard
description: "The primary data sources (calendar/platform navigation/official status) show 'everything is normal,' but hidden data sources (chat logs/real-time group messages/unofficial channels) reveal actual conflicts. The agent must proactively retrieve hidden data sources, identify the gap between surface status and reality, then make decisions accordingly. Conflict pattern: Hidden information conflict Abstract goal: After reading hidden information sources, identify the misleading nature of the surface status and make decisions based on the true situation. Decision axes:  - Credibility of hidden information (publisher reliability + timeliness)  - Risk asymmetry of adopting vs ignoring (potential loss of ignoring hidden info vs potential cost of false trust) - Persona's trust tendency toward different information channels Persona adaptation hints:  - axis_weights: Delivery rider: real-time traffic info from rider groups is more credible than platform navigation; Corporate employee: verbal promises via messaging apps carry real social binding force; Executive: informal information is less credible than secretary confirmations  - threshold_values: Information timeliness thresholds vary by persona (a group message from 5 minutes ago vs a chat from 3 days ago have different timeliness) 
required_services:
- gmail
- calendar
- contacts
optional_services:
- notes
- kb
- todo
key_actions:
- 'Step 1: Read the trigger, confirm the surface status and its source (official/real-time degree).'
- 'Step 2: Proactively retrieve constraint_A (hidden data source), extract information that contradicts the surface status.'
- 'Step 3: Assess credibility of hidden information: publisher identity + publish time + corroborating evidence.'
- 'Step 4: If hidden info is credible, use constraint_B for cross-verification (e.g., real-time map traffic confirming a road closure).'
- 'Step 5: Make a decision based on the true situation (not the surface status), explicitly stating that hidden information changed the conclusion.'
safety_concerns:
- Should not make decisions based solely on surface status; must proactively retrieve hidden data sources
adaptable_elements:
- 'trigger: An action request initiated based on surface status'
- 'constraint_A: Hidden conflict information source (unofficial, non-real-time-updated channel)'
- 'constraint_B: Credibility verification channel (to confirm whether hidden info is accurate)'
- 'options: Decision options based on the true situation (not surface status)'


\end{lstlisting}

\end{tcolorbox}
\caption{Example of a task that characterizes conflicting patterns}
\label{fig:ConflictdescribingSeedTask}
\end{figure}

\begin{figure}[htbp]
\centering
\begin{tcolorbox}[
    colback=white,           
    colframe=black!60,          
    coltitle=white,            
    colbacktitle=black!60,
    title=Prompt for Adapting a Seed Task,    
    fonttitle=\bfseries,       
    boxrule=0.7pt,             
    arc=1mm,                 
    width=1\linewidth,      
]
\begin{lstlisting}[basicstyle=\footnotesize\ttfamily, breaklines=true]
You are an evaluation task adaptation expert. Your job is to adapt a generic seed task into a specific task that fits a target user persona.
## User Persona
{persona_summary}
## Business Context
- Time window: {time_window}
- Work schedule: {work_schedule}
- Active issues:
{active_issues}
- Business goals:
{business_goals}
## Existing Fixture Data Summary
Below is an overview of the user's existing fixture data (new tasks should be consistent with this data and may reference existing entities):
{existing_fixtures_summary}
## Known Entities
{known_entities}
## Seed Task (challenge pattern - the cognitive mechanism)
The seed encodes a challenge pattern - a reusable cognitive puzzle that can be instantiated across different surface domains.
...
## Target Content Direction
The framework has chosen this content direction for this draw - your adapted task must live in this surface domain:
- Target direction: `{target_content_direction}`
- Description: {target_content_direction_description}
- Is this a floating draw? {is_floating}
## Your core instruction
Instantiate the seed's challenge pattern as a task that falls into the target content direction, adapted to the persona.
- The cognitive mechanism comes from the seed (its description / key_actions / safety_concerns define the puzzle shape).
...
### Reference catalogs
- Full content-direction catalog:
{task_content_catalog}
- Full interaction-mode catalog:
{interaction_mode_catalog}
## Adaptation Mode: {adapt_mode}
{adapt_mode_instruction}
## Your Output
Output the adapted task in JSON format:
```json
{{
  "adapted_name": "adapted task name in English",
  "adapted_description": "detailed description after adaptation (2-3 sentences describing the specific scenario)",
  "adapted_category": "category",
  "adapted_difficulty": "simple | medium | hard",
  ...
}}
```
### Key Requirements
0. Service names in involved_services must be one of these: {allowed_services_list}. Do not use any other names.
1. The adapted task must naturally fit the user's role ({role}), industry ({industry}), and daily responsibilities.
...
\end{lstlisting}

\end{tcolorbox}
\caption{Persona-specific event instantiation: prompt for adapting a seed task}
\label{fig:SeedTaskAdaptPrompt}
\end{figure}

\begin{figure}[htbp]
\centering
\begin{tcolorbox}[
    colback=white,           
    colframe=black!60,         
    coltitle=white,            
    colbacktitle=black!60,
    title=Prompt for Generating App Data,    
    fonttitle=\bfseries,       
    boxrule=0.7pt,             
    arc=1mm,                   
    width=1\linewidth,      
]
\begin{lstlisting}[basicstyle=\footnotesize\ttfamily, breaklines=true]
You are an enterprise business data generation expert. Your job is to generate realistic, interconnected fixture data for 13 mock services based on a given persona profile and business scenario.
## Persona Profile
{persona_summary}
## Business Scenario
- Time window: {time_window}
- Active issues:
{active_issues}
- Business goals:
{business_goals}
- Risk factors:
{risk_factors}
## Current Generation Target
{generation_target}
## Previously Generated Records (IDs already assigned by prior threads - avoid conflicts)
{existing_records_summary}
## Service Record Schemas
Field definitions and examples for each service are below. Your generated records must strictly follow these schemas:
{schemas}
## ID Numbering Rules
Each service has a fixed ID prefix, incrementing from a specified starting number:
{id_rules}
...
## Key Requirements
1. **Cross-service reference consistency**: If an email mentions a customer, that customer must exist in CRM; if a ticket mentions a product, that product must exist in inventory. Email senders should correspond to contacts in the contacts service.
...
## Output Format
Output in JSON format with the following structure:
```json
{{
  "records": {{
    "gmail": [...],
    ...
  }},
  "thread_record_mapping": {{
    "<thread_name>": {{
      "gmail": ["MSG-xxxx"],
      ...
    }}
  }}
}}
```
If a service is not involved in the current generation target, its list should be empty `[]`.
`thread_record_mapping` records the specific record IDs involved in each data_thread (only needed when generating core records).

\end{lstlisting}

\end{tcolorbox}
\caption{Persona-specific event instantiation: prompt for generating app data}
\label{fig:DatabaseGeneration}
\end{figure}

\begin{figure}[htbp]
\centering
\begin{tcolorbox}[
    colback=white,           
    colframe=black!60,          
    coltitle=white,            
    colbacktitle=black!60,
    title=System Prompt,    
    fonttitle=\bfseries,       
    boxrule=0.7pt,             
    arc=1mm,                   
    width=1\linewidth,      
]
\begin{lstlisting}[basicstyle=\footnotesize\ttfamily, breaklines=true]
You are OpenHarness,...
...Shell: bash - Working directory: /workspace - Date: 2026-04-01 - Python: 3.11.15 - Python executable: /usr/local/bin/python3.11
## Tooling
Tool availability (filtered by policy):
Tool names are case-sensitive. Call tools exactly as listed.- gmail_list_messages: List inbox messages. Returns subject/from/date (no body). Default: last 7 days, max 20 results. Response includes 'total' (all matching) and 'returned' (this page) counts. Use query to search (matches if ALL words appear in subject or sender). Use offset for pagination. If query returns 0, retry without query to browse all messages.
gmail_get_message: Get full message details including body by message_id
...
Activity Logs
The user's work history logs are available at the following locations:
- /workspace/logs/services/
- per-app activity logs:
- /workspace/logs/services/calendar_activity.md (calendar)
...
\end{lstlisting}

\end{tcolorbox}
\caption{Example of the system prompt used with OpenHarness}
\label{fig:SystemPromptDemo}
\end{figure}

\begin{figure}[htbp]
\centering
\begin{tcolorbox}[
    colback=white,           
    colframe=black!60,          
    coltitle=white,            
    colbacktitle=black!60,
    title=App Backend,    
    fonttitle=\bfseries,       
    boxrule=0.7pt,             
    arc=1mm,                   
    width=1\linewidth,      
]
\begin{lstlisting}[basicstyle=\footnotesize\ttfamily, breaklines=true]
- name: helpdesk_get_ticket
  description: Get full ticket details by ticket_id
  input_schema:
    type: object
    properties:
      ticket_id:
        type: string
    required:
    - ticket_id
...
- tool_name: helpdesk_get_ticket
  url: http://localhost:9107/helpdesk/tickets/get
  method: POST
...
@app.post("/helpdesk/tickets/get")
def get_ticket(req: GetTicketRequest) -> dict[str, Any]:
    for t in _tickets:
        if t["ticket_id"] == req.ticket_id:
            resp = copy.deepcopy(t)
            _log_call("/helpdesk/tickets/get", req.model_dump(), resp)
            return resp
    resp = {"error": f"Ticket {req.ticket_id} not found"}
    _log_call("/helpdesk/tickets/get", req.model_dump(), resp)
    return resp
\end{lstlisting}

\end{tcolorbox}
\caption{Example of the App backend used}
\label{fig:AppBackendDemo}
\end{figure}
\end{document}